# Unsupervised Segmentation Algorithms' Implementation in ITK for Tissue Classification via Human Head MRI Scans

Shadman Sakib
Department of Electrical and Electronic Engineering
International University of Business Agriculture and Technology
Dhaka 1230, Bangladesh
sakibshadman15@gmail.com

Md. Abu Bakr Siddique
Department of Electrical and Electronic Engineering
International University of Business Agriculture and Technology
Dhaka 1230, Bangladesh
absiddique@iubat.edu

*Abstract*— Tissue classification is one of the significant tasks in the field of biomedical image analysis. Magnetic Resonance Imaging (MRI) is of great importance in tissue classification especially in the areas of brain tissue classification which is able to recognize anatomical areas of interest such as surgical planning, monitoring therapy, clinical drug trials, image registration, stereotactic neurosurgery, radiotherapy etc. The task of this paper is to implement different unsupervised classification algorithms in ITK and perform tissue classification (white matter, gray matter, cerebrospinal fluid (CSF) and background of the human brain). For this purpose, 5 grayscale head MRI scans are provided. In order of classifying brain tissues, three algorithms are used. These are: Otsu thresholding, Bayesian classification and Bayesian classification with Gaussian smoothing. The obtained classification results are analyzed in the results and discussion section.

*Keywords—Tissue classification, Magnetic Resonance Imaging (MRI), Otsu thresholding, Bayesian classification and Bayesian classification with Gaussian smoothing*

I. INTRODUCTION

A human brain is one of the most important and complicated organ of our central nervous system. An accurate MRI classification according to the tissue type has become a demanding requirement in diagnosis, treatment planning, and cognitive neuroscience. In medical image analysis, tissue classification in MRI of the human brain is one of the key problems since the brain tissue is usually complex and the noise of the image is unavoidable. Brain MRI is a noninterfering method used for visualizing the brain soft tissues anatomy. Many techniques are used for medical imaging for visualizing the structure of the brain such as Positron Emission Tomography (PET), Computerized Tomography (CT) scan, MRI etc. Among them, MRI is the most popular and advantageous [1] because of its high-resolution images and soft-tissue contrast. Generally, the brain tissue segmentation usually classified into three primary tissue classes, named as, white matter (WM), gray matter (GM) and cerebrospinal fluid (CSF). A fourth class named as background which denotes the other areas such as skull, skin, fat, air surroundings of the head etc. The main task in tissue classification is to classify the voxels into volume measurements. Each of the pixels of an image is modeled which belongs to one of the tissue classes. However, this tissue classification has several important applications which are associated with diagnosis, surgical planning, and clinical drug trials etc. which include the study of neuro-degenerative disorders such as Alzheimer's [2], epilepsy [3], and identification of cortical surface extraction [4-8]. The exact classification of MRI according to their tissue types has become a scathing requirement in diagnosis, treatment planning, and in neuroscience. MRI segmentation is still a challenging task due to some difficulties of MRI such as intensity inhomogeneity, partial volume effects, and noise. Inhomogeneity is instigated by the radio-frequency coil in the MRI for which the image intensity values are corrupted. Partial volume effect resulted from a finite resolution of image process as well as from the difficulty of tissue edges which contain more than one tissue in a pixel causes several voxels in MRI images. Moreover, noise in the image produces dotted areas and equivocal tissue boundaries. To overcome these problems, several classification techniques are applied to brain MRI such as Gaussian mixture model (GMM) which is used for modeling the probability density function (PDF) as well as for voxel attributes, for estimating the statistical parameters efficiently, Maximum Likelihood (ML) method is used. For optimization, the Expectation Maximization (EM) algorithm is used. Moreover, Bayesian classifier, Gaussian smoothing, k-th nearest neighbors, Artificial Neural Network (ANN) are also used. In tissue intensity distribution, all MRI classification methods are sensitive to overlap. For brain tissue classification, unsupervised algorithms classification is proposed and implemented by using the ITK software manual [9].

In this paper, three methods are implemented for tissue classification. The methods are Otsu thresholding, Bayesian classification and Bayesian classification with Gaussian smoothing. The Main task is to classify the white matter, gray matter and the cerebrospinal fluid from 5 brain MRI images. Here, the most difficult part was to classify the CSF properly as a portion of this tissue had the same intensity as the background. However, all the methods demonstrated reasonable performance in case of brain tissue classification.

II. LITERATURE REVIEW

Several works have done for MRI of brain tissue classification as well as for detecting tumor based on segmentation of the image in medical image processing. Since the MRI of the human head is a challenging task, several algorithms have been developed over the years to avoid the complexity of human head tissue classification. Region-based brain MRI segmentation method [10] is proposed by Peng et al. where modified Mumford Shah's algorithm is assimilated for MRI classification into WM, GM, and CSF. Warfield et al. outlined a non-linearly deformed anatomical template where the limitations of intensity based classification are identified [11]. Van Leemput et al. proposed an automatic execution of EM statistical classification [12] which uses a probabilistic brain atlas to compel the iterative EM procedure. Moreover, for brain MRI tissue classification modified FCM algorithm is introduced by modifying each cluster and integrating spatial information which gives suitable results for MRI for distinct noise types [13]. For finding the uniform areas in an image, an edge-based segmentation strategy is outlined [14] which results very dependant on the quality of the segmentation. A Hidden Markov Chain (HMC) model is established which considers partial volume effect and inhomogeneity correction. For the segmentation of brain tumor detection a watershed algorithm is used which does not require any initialization for the segmentation of the brain MRI [15]. To solve the noise problem and computational complexity, a modified FCM algorithm is developed which estimates the bias field as well as the segmentation of MRI [16]. This allows the labeling of the voxel to be affected by the labels of its neighborhood. Furthermore, an algorithm based on spatial prior and neighboring pixels affinities were proposed for selecting a few pixels and labels them as GM, WM or CSF. To capture the partial volume effects, a linear combination model is used. Because of the modeling of partial volume effects, the results are binary instead of fractional.

### III. TISSUE CLASSIFICATION ALGORITHMS' OVERVIEW AND IMPLEMENTATION

As stated earlier, in this paper, Otsu thresholding, Bayesian classification and Bayesian classification with Gaussian smoothing are implemented on 5 different head MRI scans. The purpose is to evaluate the performances of the algorithm in case of brain tissue classification.

A Thresholding system recognizes areas based on the pixels with similar intensity values. Otsu thresholding method seeks to find the optimal threshold(s) for maximizing the inter-class variance between the classes. It is an unsupervised nonparametric method of automatic threshold selection [17]. This method uses the zeroth and the first-order cumulative moments of the gray-level histogram. Otsu's thresholding assumed to have mainly two classes of pixels which are the foreground and the background pixels. It computes the optimum threshold values to separate two classes and minimizes the variance between objects and the background and maximizes the inter-class variance. For a given threshold k, two classes prior probabilities are [18, 19]:

$$P_1(k) = \sum_{i=0}^{k-1} P_i \quad (1)$$

$$P_2(k) = \sum_{i=k}^{L-1} p_i \quad (2)$$

Therefore, the total probability is:

$$P_1(k) + P_2(k) = 1 \quad (3)$$

The class means are:

$$m_1(k) = \sum_{i=0}^{k-1} \frac{ip_i}{P_1(k)} \quad (4)$$

$$m_2(k) = \sum_{i=k}^{L-1} \frac{ip_i}{P_2(k)} \quad (5)$$

Therefore, the global intensity mean is:

$$m_G(k) = \sum_{i=0}^{L-1} ip_i \quad (6)$$

Therefore, the total mean can be expressed as:

$$m_T(k) = P_1(k) * m_1(k) + P_2(k) * m_1(k) \quad (7)$$

Now, the two classes' variances can be demonstrated by [18]:

$$\sigma_1^2(k) = \sum_{i=0}^{k-1} \frac{[i - m_1(k)]^2 * p_i}{P_1(k)} \quad (8)$$

$$\sigma_2^2(k) = \sum_{i=k}^{L-1} \frac{[i - m_2(k)]^2 * p_i}{P_2(k)} \quad (9)$$

Therefore, the total variance is:

$$\sigma_T^2(k) = \sum_{i=0}^{L-1} [i - m_T(k)]^2 * p_i \quad (10)$$

The between class variance is:

$$\sigma_B^2(k) = P_1(k) * [m_1(k) - m_T(k)]^2 + P_2(k) * [m_2(k) - m_T(k)]^2$$
, for k = 0, 1, 2,…, L-1 (11)

Hence, the class separability measure η is:

$$\eta = \frac{\sigma_B^2(k)}{\sigma_w^2(k)} \quad (12)$$

For each k, where, k=0, 1, 2,…, L-1 probabilities were calculated. Since the value for which $\sigma_B^2(k)$ is maximum, therefore, the optimum threshold is $k^*$ such that $\sigma_B^2(k^*)$ is the maximum.

For the task of this paper, we have implemented the Otsu Threshold Image Filter in ITK. It is to be mentioned that no pre-processing or post-processing is adapted for this paper. However, the outcome of the filter was reasonably good.

Bayesian classification has been widely used in various image processing applications. One of them is medical image processing. Bayesian classifier is known as a statistical classifier which estimates the class of unknown data using a probabilistic model assigned for estimating the new class of the data called Bayesian classification. In this process to segmentation, we determine the distribution of the observation noise and introduce the prior terms to classify an image into a pre-determined number of classes in the form of prior probability distribution. Bayesian classification method is initialized by generating a Gaussian mixture model for the input image. Then, at each voxel, the posterior probability for each class is computed via the Bayes rule. The voxel is assigned to the class with the highest posterior (maximum a posteriori). We will assume uniform priors. The Bayesian classifier image filter from ITK is used in this paper. The input to this filter is a vector image that represents the pixel memberships to 'n' classes. For generating the vector image, we have used the Bayesian Classifier Initialization Image Filter before implementing the Bayesian classification filter. Bayesian classification is based on the Bayes theorem. By using Bayes theorem, conditional probabilities of features can be calculated. However, Bayes theorem is expressed as [20]. By using Bayes theorem, posterior probability can be calculated as:

$$P(H|X) = \frac{P(X|H)P(H)}{P(X)} \quad (13)$$

Where X is the object, H is hypothesis whereas X belongs to a stated class C, P(H) is the prior probability, P(X|H) is the probability which the hypothesis holds given the observed data X, and P(H|X) is posterior probability which is independent of X.

The class prior probability can be approximated by:

$$P(C) = \frac{S_i}{S} \quad (14)$$

Where $S_i$ is the total number of training samples which contain class C and S is the total number of training samples.

Naïve assumption of class conditional probability is made in order to reduce the complexity of the calculation. The conditional probability is estimated using the prior probability, expressed as:

$$P(X|C) = \prod_{k=1}^{n} P(X_k|C) \quad (15)$$

Bayes' rule used for tissue classification:

$$P(C_P|I) \propto P(I|C_P)P(C_i) \quad (16)$$

Where $C_P$ is the posterior class, I is the likelihood and $C_i$ is the prior class.

Finally, the Bayesian classification with Gaussian smoothing is similar to the previous method, except that the posteriors are iteratively smoothed with a discrete Gaussian filter prior to assigning a label to the voxel. Gaussian smoothing is a 2-D convolution operator used to blur an image and reduces image noise and details. It is a non-parametric classification method and assumes few prior distributions on the underlying probability densities that assurances some smoothing properties. This is accomplished by taking the smoothness prior to the account while factoring in the observed classification of training data. The 2-D convolution is implemented by first convolving with a 1-D Gaussian in the direction of x and then convolving with another Gaussian in the y-direction. Gaussian smoothing is also known as Gaussian blur. The degree of smoothing is determined by the standard deviation. However, Gaussian smoothing is advantageous because it reduces the size of an image by downsampling it. It is also used in edge detection purposes to reduce the noise within the image. In one dimension, the Gaussian function is:

$$G(x) = \frac{1}{\sqrt{2\pi\sigma^2}} * \exp\left(-\frac{x^2}{2\sigma^2}\right) \quad (17)$$

In two dimensions:

$$G(x,y) = \frac{1}{2\pi\sigma^2} * \exp\left(-\frac{x^2 + y^2}{2\sigma^2}\right) \quad (18)$$

Where σ is the standard deviation of the distribution, x is the distance from the origin in the horizontal axis and y is the distance from the origin in the vertical axis.

## IV. RESULTS AND DISCUSSION

At the beginning of the implementation, we were trying to implement the classification algorithms to the MRI images by defining the number of classes as 3. However, the outcomes of the implementations were not satisfactory. In these cases, though the algorithms were able to differentiate the white matter and the gray matter, the classification of the cerebrospinal fluid (CSF) was not quite good. Then we have planned to try different class numbers. This resulted in the better classification of CSF portion in the MRI.

TABLE I. BRAIN TISSUE CLASSIFICATION RESULT FOR DIFFERENT CLASSIFICATION METHODS FOR 1ST TWO MRI SCANS

| Input Image | Otsu thresholding Result | Bayesian classification Result | Bayesian classification with Gaussian smoothing Result |
|---|---|---|---|
| 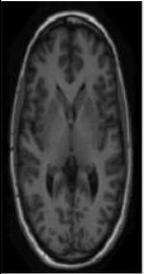 | 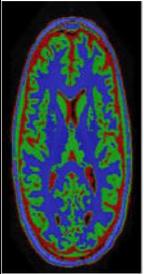 | 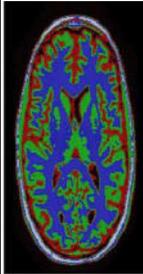 | 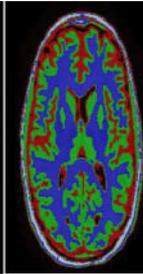 |
| 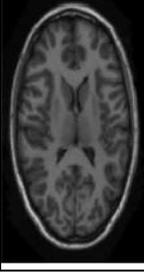 | 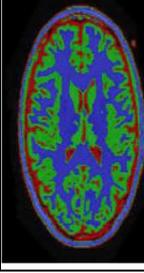 | 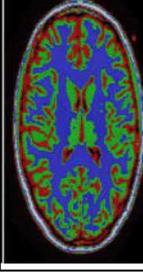 | 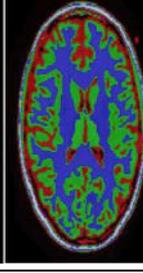 |

In the results of the algorithms, the connecting edges of different tissues were not smooth. Here, the smoothing filter demonstrates better results. We have tried the number of iteration as 5, 7 and 10. However, according to our observation, the better result in this case is found for 5 iterations. The time step and the conductance parameter are set to be 0.05 and 3 respectively.

Comparing the algorithms used in this paper, it is evident that all the methods are performing reasonably well for the provided data. It is to be noted that, neither pre-processing nor post processing of the data was adapted. Therefore, the algorithms, in some cases, classified a portion of the CSF wrong. In comparison, the Bayesian classification with Gaussian smoothing demonstrated better result in case of identifying the CSF in between the gray matter and the skull. On the contrary, Otsu demonstrated better performance while classifying the CSF region in the center of the brain. However, the implementation of the Otsu classifier was simple and less computationally expensive.

We were pretty satisfied with the results, however, the filters, in some cases, was classifying a small portion of the CSF as the background. According to our observation, this misclassification is due to the similar intensity of the background and a small portion of the CSF. This could have been corrected by applying some preprocessing or post processing approaches.

We have learned to implement Otsu thresholding, Bayesian classification and Bayesian classification with Gaussian smoothing for brain tissue classification. Though there are some limitations in the classification algorithms, the performance can be improved by augmenting other image processing techniques to those methods. We have also learned the way of conversion of an image into a vector image which was implemented in Bayesian classification algorithm. Tissue classification is a tricky task in the case of MRI images. It is sometimes needed to indicate a higher number of classes in the algorithm to produce a better result. Overall, the results were reasonably good.

TABLE II.   BRAIN TISSUE CLASSIFICATION RESULT FOR DIFFERENT CLASSIFICATION METHODS FOR NEXT THREE MRI SCANS

| Input Image | Otsu thresholding Result | Bayesian classification Result | Bayesian classification with Gaussian smoothing Result |
|---|---|---|---|
| 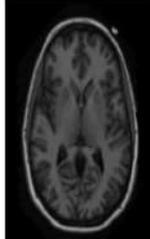 | 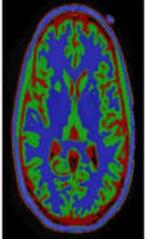 | 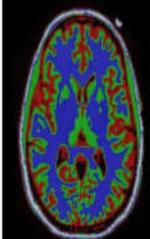 | 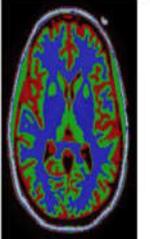 |
| 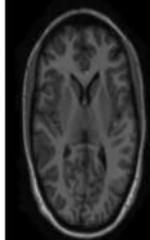 | 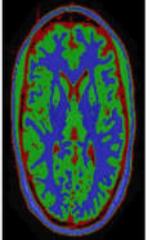 | 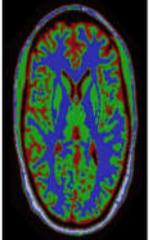 | 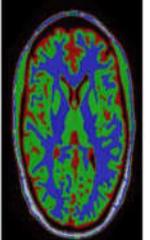 |
| 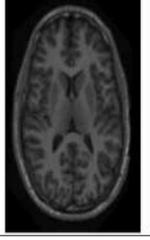 | 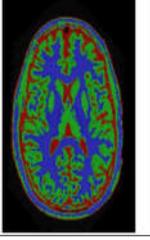 | 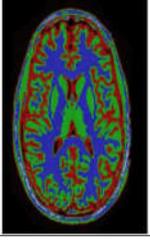 | 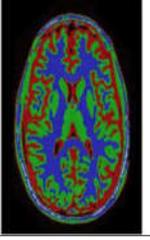 |

## V. CONCLUSION

The performance of different algorithms for classifying the gray matter, white matter and the CSF was very interesting. It was needed to indicate more number of classes for getting the desirable result. All the classifying algorithms used in this paper showed great performance in classifying the gray matter and white matter. The classification of the CSF portion was a bit tougher. The Ostu classifier identified the CSF located in the center of the brain well. On the other hand, Bayesian classifier and the Bayesian classifier with Gaussian smoothing could correctly identify the CSF located in between the gray matter and the skull.